\documentclass{article}

\usepackage{arxiv}

\usepackage[utf8]{inputenc} 
\usepackage[T1]{fontenc} 
\usepackage{hyperref}  
\usepackage{url}   
\usepackage{booktabs}  
\usepackage{amsfonts}  
\usepackage{nicefrac}  
\usepackage{microtype}  
\usepackage{graphicx}
\usepackage{caption}
\usepackage{subcaption}
\usepackage{xspace}
\usepackage{geometry}
\usepackage{amsmath,amssymb,amsfonts,dsfont}
\usepackage{breqn}
\usepackage{lmodern}
\usepackage[lined,  ruled, linesnumbered]{algorithm2e}
\usepackage{xcolor}

\title{Learning  Vehicle Routing Problems using Policy Optimisation}
\author{Nasrin Sultana, Jeffrey Chan, A. K. Qin, Tabinda Sarwar}

\begin{document}
\maketitle

\begin{abstract}

Deep reinforcement learning (DRL) has been used to learn effective heuristics for solving complex combinatorial optimisation problem via policy networks and have demonstrated promising performance. Existing works have focused on solving (vehicle) routing problems as they have a nice balance between non-triviality and difficulty. State-of-the-art approaches learn a policy using reinforcement learning, and the learnt policy acts as a pseudo solver. These approaches have demonstrated good performance in some cases, but given the large search space typical combinatorial/routing problem, they can converge too quickly to poor policy. To prevent this, in this paper, we propose an approach name entropy regularised reinforcement learning (ERRL) that supports exploration by providing more stochastic policies, which tends to improve optimisation. Empirically, the low variance ERRL offers RL training fast and stable. We also introduce a combination of local search operators during test time, which significantly improves solution and complement ERRL. We qualitatively demonstrate that for vehicle routing problems, a policy with higher entropy can make the optimisation landscape smooth which makes it easier to optimise. The quantitative evaluation shows that the performance of the model is comparable with the state-of-the-art variants. In our evaluation, we experimentally illustrate that the model produces state-of-the-art performance on variants of Vehicle Routing problems such as Capacitated Vehicle Routing Problem (CVRP), Multiple Routing with Fixed Fleet Problems (MRPFF) and Travelling Salesman problem.

\end{abstract}


\section{Introduction}

Combinatorial Optimisation (CO) Problems are hard as it involves finding the optimal solution under various constraints. A conventional approach to solve these problems involves modelling the problem into a mathematical objective, selection of an appropriate solver and then optimising its parameters for the problem instance at hand. While this approach has been successful, it requires high levels of optimisation expertise and domain knowledge, limiting their widespread usage. Also, the selection of solver and its optimal parameters varies for different problem instances when the problem instance changes, often the process of searching appropriate solver and parameters are restarted. This has raised interest in the level of generalisation at which optimisation operate.

In recent years, there is increasing interest in considering combinatorial optimisation as a learning problem,~(\cite{bello2016neural};~\cite{khalil2017learning}), where optimisation instances, and their solutions, are used as training instances~(\cite{vinyals2015pointer}). Thereby, the resulting learnt model is then considered as a general solver. The models typically involve deep neural networks (DNNs)~(\cite{burke2emerging}), and recent state-the-art approaches take a reinforcement learning strategy for cases where supervised approaches/solutions are not plausible~(\cite{vinyals2015pointer}). The learnt policy is similar to a solver~(\cite{kool2018attention}). In particular, deep reinforcement learning (DRL) and policy gradients can successfully finding close to optimal solution to the problems such as TSP \cite{bello2016neural}, \cite{deudon2018learning}, \cite{ma2019combinatorial} CVRP~(\cite{nazari2018reinforcement}, ~(\cite{kool2018attention}), 0-1 Knapsack~\cite{bello2016neural}. All the problems we discussed in Appendix~\ref{apd:problems}.

The previously mentioned state-of-the-art~(\cite{nazari2018reinforcement} and~\cite{kool2018attention}) uses standard policy gradient-based approaches (PGM) build upon the REINFORCE algorithm~(\cite{williams1992simple}). For standard reinforcement learning (RL) problems, search space (landscape) is smaller, the search space is smooth, and optimisation is not difficult. However, in the case of combinatorial optimisation ((class of CO) problems such as VRP, TSP, search space may not be infinite, but the search space of the CO is difficult \cite{weinberger1990correlated}. Therefore, for combinatorial optimisation problems, we need an effective method for moving through the search space. The issue with the standard policy gradient methods apply to CO Problem is, fluctuating landscape of the CO Problem which may not estimate gradient properly (two nearby samples can have very different gradients) which make the problem difficult to optimise. Because when we take the average which may have variance in the gradient. Additionally, in many cases in RL, a situation can occur if the agent discovers a strategy following a policy that achieved a reward which is better than when the agent started first, but the strategy of the agent following can be difficult to find the optimal solution and tend to take a single move over and over. So as agent progressing learning, it is average move distribution will be closer to prediction with either single move or multiple moves. So it is unlikely to explore different actions.  
As a consequence, instead of using standard PGM, our line of research devoted to how can we explore the search space effectively and efficiently to improve the solution of a combinatorial optimisation problem?  When we look for the best possible answer to the question, a perfectly logical solution to use exploration strategy for better exploration in the search space.  As a result, we need a method that can able to explore the search space effectively. However, this is not the case in previous learning methods. To solve this, we add an entropy of the policy with the RL objective as in \cite{ziebart2010modeling}. The goal is to find the probability distribution that has the highest entropy, which states that it is one of the best representation of the current state. Using exploration strategy in RL can be used where the neural network is used as function approximation. 

In the line of research, we contribute to the deep learning community for COP by introducing entropy regularised term in recent policy gradient-based methods (ERRL). ERRL offers to improve policy optimisation \cite{ahmed2018understanding}. It is believed that entropy regularisation, assist with exploration by encouraging the selection of more stochastic policies~(\cite{ahmed2018understanding}). As a consequence, in this work, we analyse this claim a policy with higher entropy can make the changes the optimisation landscape and maintains exploration to discourage early convergence. To the best of our knowledge, the entropy regularised term has not been studied or used in existing learning to (combinatorial) optimise literature. ERRL can be integrated with any existing policy gradient approaches that use parameterised functions to approximate policies; hence we applied entropy technique to the state-of-the-art methods~(\cite{nazari2018reinforcement});~(\cite{kool2018attention}). We demonstrated the effectiveness of ERRL on three categories of routing problems. The goal of this work is not to outperform all the existing state-of-the-art VRP learning algorithm from every aspect but to provide direction in the study of the RL approach to encourage exploration to fundamental routing problems, considering the before-mentioned difficulties. 

The main contributions are as follows:

\begin{itemize}
\item We proposed an approach using entropy regularised term that can solve route optimisation problems. We devise a new exploration-based and low-variance method for policy gradient method because this baseline assists to select a more stochastic policy. The proposed method is verified on multiple types of routing problems, i.e., Capacitated vehicle routing problem (CVRP) and multiple routing with fixed fleet problems (MRPFF) and Travelling salesman problems (TSP). 

\item  The generality of the proposed scheme is validated with different approaches and evaluating the resultant method on various problem sizes (and even at high problem dimensionality of 100) to achieve outstanding performance, better than the state-of-the-art in terms of accuracy and time-efficiency. 

\item Another contribution we use a local search algorithm 2-opt~(\cite{aarts2003local}). This hybrid approach is an example of combining learned and traditional heuristics to improve the solution. In this work, we analyse existing inference techniques to show the impact of the post-processing techniques in the solution quality.

\end{itemize}

\section{Related Work}\label{sec:related}

In recent years, the line of research has many ways to solve COP using deep learning paradigm.  Many methods have been developed to tackle combinatorial optimisation problems.  Traditional heuristics for routing problems can be categorised as construction and improvement heuristics(\cite{toth2014vehicle}).

\textbf{Construction methods (Supervised).} Recent advances in the neural networks include the design of a new model architecture called Pointer Network(PN)~(\cite{vinyals2015pointer}). In Pointer network~(\cite{vinyals2015pointer}) learns to solve combinatorial optimisation problems where encoder (RNN) converts the input sequence that is fed to the decoder (RNN). They use attention on the input and train this model in a supervised setting to solve the Euclidean TSP instances. The goal is to use the Pointer Network architecture to find close to optimal tours from ground truth optimal (or heuristic) solutions for Traveling Salesman Problem (TSP). 

In~(\cite{joshi2019efficient}) takes a graph as an input and extracts features from its nodes and edges. Their model can be considered as a  stack of several graph convolutional layers. The output of the neural network is an edge adjacency matrix representing the probabilities of edges occurring on the TSP tour. The edge predictions, forming a heat-map, are transformed into a valid tour.  They trained their model as a supervised manner using pairs of problem instances and optimal solutions. 

\textbf{Constructive methods (Deep RL).} Despite this promising early application, reinforcement learning becomes a compelling choice to the prospect of learning to optimise as it does not require a set of pre-solved solutions for training. ~(\cite{bello2016neural}) first propose a reinforcement learning approach, in which a pointer network is trained using an actor-critic reinforcement learning strategy to generate solutions for artificial planar TSP instances. They address this issue by designing a neural combinatorial optimisation framework that uses reinforcement learning to optimise policy. S2VDQN~(\cite{khalil2017learning}) solves optimisation problems using a graph embedding structure and a deep Q-learning algorithm.

Recently, many deep learning-based approaches exist, however only a few learning-based approaches propose a solution to the VRP~(\cite{nazari2018reinforcement},~\cite{kool2018attention}). Recent approaches applied a deep RL model that generate solutions sequentially one node at a time. The constructive heuristics,~(\cite{nazari2018reinforcement}), a model was proposed that uses a recurrent neural network (RNN) decoder and an attention mechanism to build solutions for the CVRP and the SDVRP, train the model using policy gradient methods (actor-critic approach) similar to~(\cite{bello2016neural}). Solution searching techniques used a beam search with a beam-width of up to 10. A graph attention network similar to~(\cite{deudon2018learning}) is used in~(\cite{kool2018attention}) and generate solutions for different routing problems trained via RL, including TSP and CVRP. They train their model using policy gradient RL with a baseline based on a deterministic greedy rollout. 

Our work can be classified as constructive method for solving CO problems, our method differs from previous work. First, we applied the entropy maximisation techniques to carried out by adding an entropy regularisation term to the objective function of RL to prevent premature convergence of the policy and change in the gradient. Second, we combine classical heuristics with improving the solution quality further. In order to promote the idea of exploring the search space, we need an effective exploration-based algorithm. The fundamental property of regularised term distinguishes our approach from rest. Our approach Entropy Regularised RL(ERRL) can be summarised as follows: instead of learning deterministically and making a decision at an early stage, we demonstrate that stochastic state-space models can be learned effectively with a well-designed network encouraging exploration. 

\textbf{Improvement methods.} Other work focuses on iteratively improving heuristics, for example,~(\cite{chen2019learning}) propose an RL based improvement approach that iteratively chooses a region of a graph representation of the problem and then selects and applies established local heuristics. A perturbation operator further improved this approach~(\cite{lu2019learning}). 

After training the neural network, some existing techniques can be applied to improve the quality of solutions it generates, for example in bello et al.~\cite{bello2016neural}, active search optimises the policy on a single test instance.  In bello et al.~\cite{bello2016neural};~\cite{kool2018attention} used Sampling method that select the best solution among the multiple solution candidates. Beam search is another widely used technique \cite{vinyals2015pointer} uses to improve the efficiency of sampling. To further enhance the quality of the solution in \cite{deudon2018learning} one popular local search operator used. Following \cite{deudon2018learning}, in ERRL, we combine classical 2-opt local search operator to process the solution further to improve. In addition, in this work, we combine many inference techniques with our ERRL method to show the importance of searching in ML-based approaches to combinatorial optimisation \cite{franccois2019evaluate}.

\section{Motivation}
In recent works, PGM use to solve CO problem. The key idea in policy optimisation (PGM) is to learn parameters,~$\theta$  of a policy,~$\pi_{\theta}(a|s)$, $s \in S, a \in A$. Here a is the action and s is state. The policy gradient method~(\cite{sutton2000policy}) stated the gradient as:

\begin{equation}
   J_{ER} (\theta) =   \sum_{s\in S} d^{\pi_{\theta}}(s) \sum_{a\in A}  \pi_{\theta} (a|s) Q^{\pi_{\theta}}(s,a)
\label{equation:theta1} 
\end{equation}

where $d^\pi$ is the stationary distribution of states and~$Q^{\pi}_{\theta}(s_t,a_t)$ is the expected discounted sum of rewards starting state~$s$, taking action~$a$ and sampling actions according to the policy,~$a\sim \pi(. |s_{t})$. The ~$Q^{\pi}_{\theta} (s,a)$), (Monte Carlo estimation) is the value function pair following a policy $\pi$. Here we are interested in finding parameter $\theta$ that maximises the objective function $J_{ER}$ (the goal is to maximise the discounted cumulative rewards). The equation helps to find a policy with the highest expected reward from the agent's action. However, many issues are encountered using current PGM approaches in combinatorial optimisation problems as combinatorial optimisation is not easy. One approach to characterising the degree of difficulty of an optimisation problem is its search space. The search space is also known as its landscape, and solutions to the optimisation are points on this landscape. It is difficult to have access to the exact transition and reward dynamics~\cite{ahmed2018understanding}. Therefore, the gradient of~$J_{ER}$ given by the policy gradient theorem in Equation~\ref{equation:theta1} cannot be evaluated directly. The Equation~\ref{equation:theta1} allowing us to estimate $\nabla J_{ER}$ using Monte-Carlo samples, where each trajectory is defined as a sequence. In a previous learning-based model, never consider an agent that have prior knowledge of previous states, so it is perfectly logical to think of an agent that needs to have experience of previous states of data to achieve maximum reward (encourage exploration). In consequence, need to change in the RL objective to solve COP so that model can achieve the highest expected reward with low variance.


\section{Entropy Regularised Reinforcement Learning(ERRL)}
\subsection{Encourage Exploration}
In our model we are given a problem s (a set of nodes $(n_1,\cdots,n_m)$) and a policy $\pi$ which is parameterise by ${\theta}$. The policy is trainable and can be produce a valid solution to the problem. The solution is as ($L= {a_1,\cdots, a_i}$), where the ith action $a_i$ can choose a node in the tour(solution). The neural network generated the solution one node at a time in a auto-regressive manner following the policy(stochastically), $\pi_{t} = P_{\theta} (a_t | s)$, where, $t=1$ and s is the problem instance (define as state). 

\begin{figure}[ht]
\begin{center}
\includegraphics[width=0.75\textwidth]{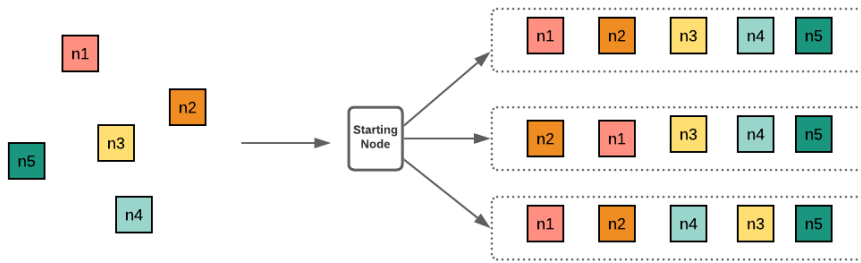}
\caption{A problem can has multiple solutions, here left figure is the 5 nodes TSP problems, in the right figure illustrates a problem could have multiple solutions that we can consider as feasible solutions.}\label{fig:policy}
\end{center}
\end{figure}

\begin{figure}[ht]
\begin{center}
\includegraphics[width=0.6\textwidth]{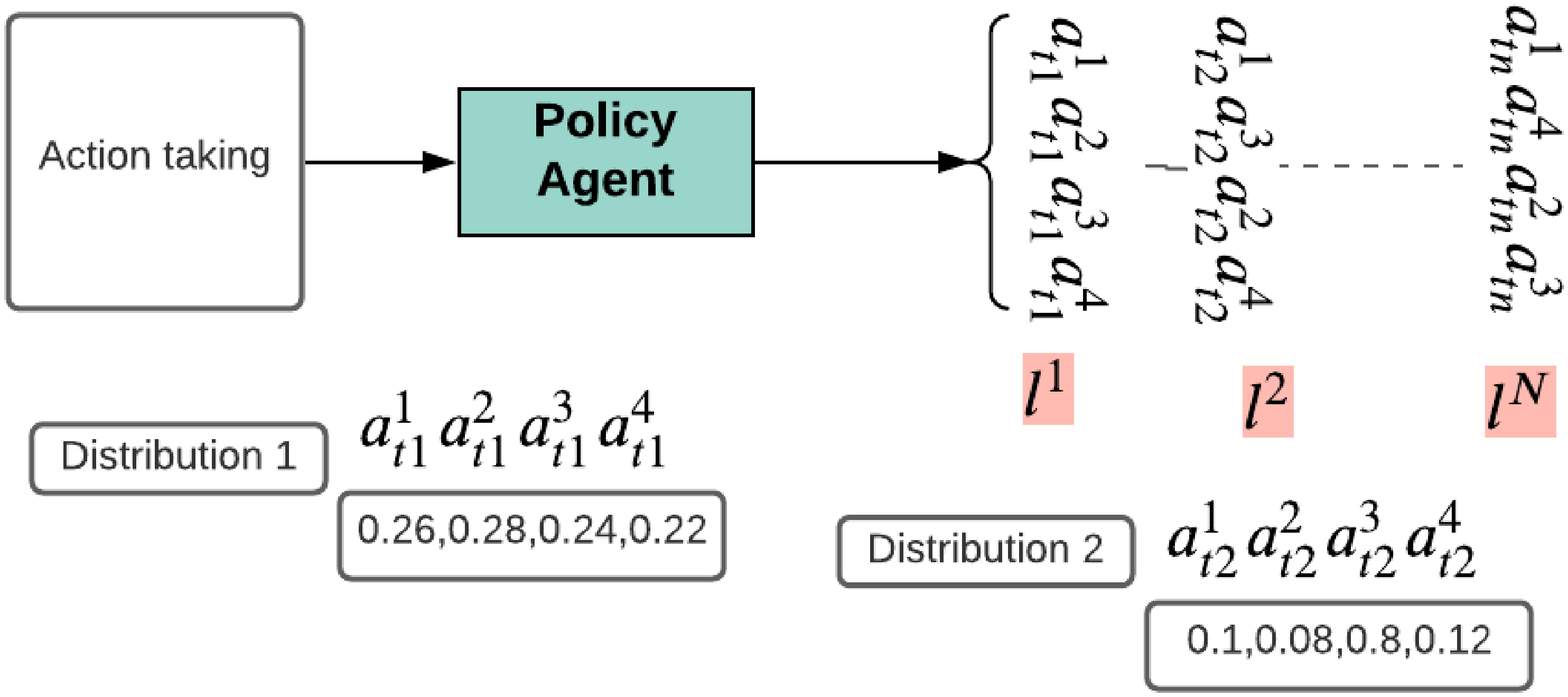}
\caption{ERRL method taking diverse action in parallel encouraging exploration}\label{fig:errl}
\end{center}
\end{figure}

In ERRL methods, we offer exploration strategy in the CO problems. Our ERRL model starts with a node for exploration shown in Figure~\ref{fig:errl}. The network samples N trajectories (${{l^1,l^2,\cdots,l^N}}$) using the Monte Carlo method, where each trajectory is defined as $l^i$. In Figure~\ref{fig:errl} presented how ERRL model learns exactly previous stated data (in our case, these stated previous data is the experience of the agent.), Figure~\ref{fig:errl} shows the two distributions of the Q values can be represent as for $(distribution1 = a_0 0.26, a_1 0.28, a_2 0.24, a_3 0.220)$ and $(distribution2 = a_0 0.1, a_1 0.08, a_2 0.8, a_3 0.12)$, for the first distribution all of the probabilities are low and from the second distribution $a_2$ has a high probability and other actions has lower probability. It had happened some cases that the agent will use the action in the future, as it already achieves some positive reward when it started.  Therefore, the agent does not tend to explore in this context; however, there could exist another action that could have a much higher reward. Consequently, the agent will never try to explore instead will exploit what it has already learned. This means the agent can get stuck in a local optimum because of not exploring the behaviour of other actions and never finding the global optimum. Hence adding entropy, we encouraged exploration and avoid getting stuck in local optima. The use of Entropy in RL works as when the agent is progressing learning the policy, according to the agent action model returns a more positive reward for the state.  The entropy augmented in the policy in the conventional RL objective following~\cite{ziebart2010modeling} formalised in Equation~\ref{equation:entropy}. 

In this work, entropy maximisation is typically carried out by adding an entropy regularisation term to the objective function of RL. Therefore, when all actions are equally good in ERRL the entropy regularisation~\cite{ziebart2010modeling} improve policy optimisation in reinforcement learning maximising reward to improve exploration. By augmenting entropy regularisation with the reward that helps to get more reward proportion to the entropy of the~$\pi$ in the following equation:
\begin{equation}
\begin{aligned}
\pi^* =  \operatorname*{argmax}\limits_{\pi}\sum_{t=1}^{T} \operatorname*{E}\limits_{(s^{t},a^{t)} \sim P_{\pi}} [ R(s_t, a_{t}) +\alpha  H (\pi_{\theta} (. |s_{t})) ] ,
\label{equation:entropy}
\end{aligned} 
\end{equation} 
It supports to help with exploration by encouraging the selection of more stochastic policies. Where~$\pi$ is a policy, ${\pi^*}$ is the optimal policy, $T$ is the number of time-steps, state~$s\in S$ is the state at time-step~$t$,~$a\in A$ is the action at time-step~$t$,~$P_{\pi}$ is the distribution of trajectories induced by policy~$\pi$. $H (\pi_{\theta} (. |s_{t})$ is the entropy of the policy~$\pi$ at state $s_t$ and is calculated as~$H (\pi_{\theta} (. |s_{t})= -log(\pi_{\theta} (. |s_{t})$, where H is the entropy and~$\alpha$ controls the strength of the entropy regularisation term. $\alpha$ is a temperature parameter that controls the trade-off between optimising for the reward and for the entropy of the policy, the entropy bonuses play an important role in reward ($\alpha$ value discusses in Section~\ref{apd:dataset}). So with the adding entropy to maximise objective will help agent to explore mode and have knowledge of the best representation of the state likely to have a probability distribution with the highest entropy, which means agent can have to the experience of the stated prior data. 

This incorporated entropy term, defined over the outputs of the policy network, into the loss function of the policy network, and the policy exploration can be supported to maximising the reward. In other words, add an entropy term into the loss function, encouraging the policy to take diverse actions). The entropy $H(\pi(a_t|s))$ used into the loss function to encourage the policy $\pi_t = \theta(a_t|s)$.  
We use the entropy regularisation~\cite{ziebart2010modeling} term with the current approaches~\cite{nazari2018reinforcement} and ~\cite{kool2018attention}, by augmenting the rewards with an entropy term, $H(\pi(. |s_{t}))= E_{a\sim\pi(. |s_{t})} [-log \pi(a|s_t)]$. This entropy regularisation term is weighed by~$\alpha$. It helps exploration by encouraging the selection of more stochastic policies (over deterministic ones), and premature convergence can also be prevented and learning is stable and sample efficient we demonstrates in the experiment section.

In Equation~\ref{equation:theta1},~$Q^{\pi_\theta}(s,a)$ changed to ~$Q^{\alpha,\pi_\theta}(s,a)$ is the expected discounted sum of entropy-augmented rewards.~$Q^{\alpha,\pi_\theta}(s,a)$ can be estimated by executing~$\pi_{\theta}$ in the environment~\cite{williams1992simple}. Because of this term we can have slightly different value function in the settings which change the value because of the included entropy at every time step. This term encourages the policy to assign an equal probability of action that has the same as the total expected reward or nearly equal to the reward by exploring the search space. Also, helps agent not to select a particular action repeatedly that could exploit inconsistency in the approximation of~$Q^{\pi_\theta}(s,a)$ which that can suffer high variance, instead can explicitly encourage exploration. The expected discounted sum of reward $Q^{\pi}(s_t,a_t)$ depends on the policy $\pi_{\theta}$, so any change in the policy will be reflected in both the objective and gradient. 


For all our experiment, use the policy network introduces by Kool et al.~\cite{kool2018attention} and Nazari et al. ~\cite{nazari2018reinforcement} named as ERRL1 and ERRL2 respectively. The Kool et al. \cite{kool2018attention} methods employed attention mechanisms, and the model has an encoder and decoder neural network. The associations between the nodes of the graph are captured by a multi-head attention mechanism, using a rollout baseline in the REINFORCE algorithm, which is a stack of layered encoder. We also use the same attention decoder from Kool et al. \cite{kool2018attention}, the decoder generates a solution sequence autoregressively), which starts from a random node and outputs a probability distribution over its neighbours at each step. The Nazari et al. \cite{nazari2018reinforcement} methods also contain an encoder and decoder.  Due to space limitations, we describe the policy networks briefly for Kool et al. ~\cite{kool2018attention} in Appendix~\ref{apd:am}. The ERRL2 model neural architecture introduce by Nazari et al.~\cite{nazari2018reinforcement} described in ~\ref{apd:nazari}. Algorithm ~\ref{algo:ERRam} presents the \textbf{ERRL1} training (follow the same training setup as \cite{kool2018attention} and Algorithm~\ref{algo:ERRLLNet}, presents the \textbf{ERRL2} training (follow the same training setup as \cite{nazari2018reinforcement}).

\textbf{Training [ERRL1]}\label{apd:errl1}

ERRL1 is based on Reinforce with a simple baseline to train the model. We used proposed attention model~(\cite{kool2018attention}) and trained their model using reinforce with baseline policy gradient method. We trained their model using our Entropy Regularised Reinforcement Learning, name ERRL1~\cite{kool2018attention}. The problem instance defines as $s$ is a graph, and the model is considered as Graph Attention Network~(\cite{velivckovic2017graph}). Attention based model defines a stochastic policy $p(\pi|s)$ for selecting solution $\pi$ given the problem instance $s$, parameterised by $\theta$. As solution define as $\pi$ and network samples N solution trajectories~{${l^1,l^2,\cdots,l^n}$}, we calculate the total reward $R(l^i)$ of each solution $l^i$. We use gradient ascent with an approximation to maximise the expected return J:

\begin{equation}
\begin{matrix}
     \nabla_{\theta}J(\theta) \approx \dfrac{1}{N} \sum_{i=1}^{N} (R(l^i) - b^i(s)))\nabla_{\theta} log P_{\theta}(l^i|s)
   \label{equation:pg}
\end{matrix}
\end{equation}

In order to train our ERRL1 model, we optimise the objective by using the REINFORCE~(\cite{williams1992simple}) gradient estimator with augmented entropy term along with baseline. To encourage exploration and avoid premature convergence to a sub-optimal policy~\cite{mnih2016asynchronous}, we add an entropy bonus.

\begin{equation}
\begin{matrix}
     \nabla_{\theta}J(\theta) \approx \dfrac{1}{N} \sum_{i=1}^{N} (R(l^i) - b^i(s)))\nabla_{\theta} log P_{\theta}(l^i|s)+ \alpha H(\pi_\theta(.|s_t)))
   \label{equation:pg1}
\end{matrix}
\end{equation}

Here, reinforcement learning with baseline learns both $R(l^i)$ and the ${b^i(s)}$ used as a baseline. The average returns serve as a baseline with included entropy bonuses. Here, we use entropy term to prevent premature convergence, results in a slightly different gradient with changing value function. This will result in larger entropy, which means the policy will be more stochastic.

\begin{algorithm}
\caption{ERRL1}\label{algo:ERRam}
\footnotesize
\SetKwFunction{This}{this}
Input: Training Set S, number of nodes per sample N, steps per epoch T, batch size B

Initialise Policy network parameters as $\theta$

    \For{$step = 1$, $\cdots$ ,$T$}{

         $s_i \leftarrow$ randomInstance(S) $\forall_i \in$ \{$1$ $\cdots$  $B$\}
         
         ${\beta^1{_i},\beta^2{_i},\cdots,\beta^N{_i}} \leftarrow$ sampling($s_i$) $\forall_i \in$ \{$1$ $\cdots$ $B$\}
         
         $l^j_{i} \leftarrow $ rollout ($s_i$, $\beta^j_{i}, \pi_{\theta} $)
         $\forall_i \in$ \{$1$ $\cdots$ $B$\}  $\forall_j \in$ \{$1$ $\cdots$ $N$\}

        $\nabla_{\theta}J(\theta)  \leftarrow \dfrac{1}{NB} \sum_{i=1}^{B} \sum_{j=1}^{N}(R(l^i) - b^i(s)))\nabla_{\theta} log P_{\theta}(l^i_{j}|s)+ \alpha H(\pi_\theta(.|s_t)))$
         
         
         $\theta \leftarrow Adam (\theta, \nabla_{\theta}J(\theta))$
}
\end{algorithm}

A good baseline $b^i(s)$ reduces gradient variance and therefore increases the speed of learning. After generating solution trajectories (${l^1,l^2,\cdots,l^n}$), we used the greedy-rollout baseline scheme, each sample-rollout assessed independently. With the changed baseline, now, each trajectory competes with N-1 other trajectories where the network will not select two similar trajectories. With the increased number of heterogeneous trajectories all contributing to setting the baseline at the right level, premature converge to a suboptimal policy is not encouraged instead converge to explored policy. Afterwards, similar to~(\cite{kool2018attention}) we updates via ADAM (adaptive moment estimation)~(\cite{kingma2014adam}) combine the previous objectives via SGD (stochastic gradient descent). Details are given in Algorithm~\ref{algo:ERRam}.

\section{{Experiments}}\label{sec:exp} 
All of our ERRl experiment use the PGM. We emphasize that ERRL can be applied to any PGM, to support the claim we applied ERRL to existing two algorithms introduce by Kool et al.~\cite{kool2018attention} and Nazari et al.~\cite{nazari2018reinforcement}. 

\textbf{Problem setting, dataset}

In this work, we implemented the datasets described by~(\cite{nazari2018reinforcement}) for TSP and CVRP for ERRL1 and ERRL2. For MRPFF we implemented dataset, where we need to find the shortest path connecting all N $(n_1,\cdots,n_i)$ nodes, where the distance between two nodes is 2D Euclidean distance. The location of each node is sampled randomly from the unit square, in this problem vehicle does not need to full fill any customer demand but optimise multiple routes.

\textbf{Training details ERRL1}

We used the same architecture settings as~\cite{kool2018attention} throughout all the experiments and initialize parameters Uniformly like \cite{kool2018attention}, policy gradients are averaged from a batch of 128 instances. Adam optimizer~\cite{kingma2014adam} is used with a learning rate  0.0001 and a weight decay (L2 regularisation). To keep the training condition simple and identical for all experiments we have not applied a decaying learning rate, although we recommend a fine-tuned decaying learning rate in practice for faster convergence. Every epoch we process 2500 batches of 512 instances generated randomly on the fly. Training time varies with the size of the problem. In Figure~\ref{fig:smalltsp} illustrates most of the learning is already completed by 200 epochs. In the experiment, we manually set entropy values for all the problems. We evaluated the results for other values results presented in Appendix~\ref{section:parameter}, but the best performance was achieved with $\alpha=0.3$ with learning rate  0.0001.
\begin{table*}[h!]
\centering
\caption{Average tour length, gap percentage and the average solution time.  * indicate values reported from~(\cite{kool2018attention}). We reporter Chen and Tian~\cite{chen2019learning} and L2I~\cite{lu2019learning} results from their paper. Nazari et al.~\cite{nazari2018reinforcement} and Kool et al.~\cite{kool2018attention} we implemented and reported the results using their publicly available codes. The gap percentages reported with respect to optimal value. The running time reported in minutes (m) and seconds (s).}
\scriptsize
\begin{tabular}{|l|lll|lll|lll|}\hline
Method & \multicolumn{3}{|c|}{CVRP20} & \multicolumn{3}{|c|}{CVRP50}& \multicolumn{3}{|c|}{CVRP100} \\ \hline
\textbf{CVRP}    & TourL  & Gap(\%) & Time(s) & TourL & Gap(\%) & Time(s) & TourL & Gap(\%) & Time(s) \\\hline\hline 
LKH3 & 6.14  &  0.00 & 28(m)  & 10.39 & 0.00\% & 112(m) &  15.67 &0.00 & 211(m) \\ 
Random CW*& 6.81  & 10.91  &   & 12.25&  17.90 &  & 18.96 & 20.99 &  \\ 
Random Sweep*& 7.01 &  14.16  &  & 12.96 & 24.73 &  & 20.33 & 29.73 &\\ 
Or-tools* &  6.43& 4.73   & &11.43 & 10.00  & & 17.16& 9.50 &  \\\hline
\textbf{Improvement Models} &  &  & &  &  &  &  &  &  \\\hline
Chen and Tian & 6.16& 0.3& -  & 10.51& 1.15 & - &  16.10  &2.72 & - \\
L2I & 6.12 & - & 12(m)  & 10.35 & - & 17(m) & 15.57 & - &  24(m) \\\hline
\textbf{Constructive Models} &  &  & &  &  &  &  &  &  \\\hline

Kool\cite{kool2018attention}& 6.67 & 8.63& 0.01  & 11.00  & 5.87& 0.02  & {16.99} &8.42&  0.07\\
Nazari \cite{nazari2018reinforcement}& 7.07& 15.14& 6.41  & 11.95& 15.01 & 19 &  17.89 &14.16 & 42 \\\hline
\textbf{Ours(Constructive)} &  &  & &  &  &    \\\hline

\textbf{ERRL1}  & {6.34} & 3.25 & 0.01  & {10.77} & 3.65 & 0.02  & {16.38} & 4.53 &  0.06 \\ 
\textbf{ERRL2} & {6.67} & 8.63  & 5.4  & {11.01} & 2.63 & 14  & {17.23} & 9.95 & 33 \\\hline
\textbf{ERRL1(2opt)} & \textbf{6.13} & - & 2.49(m)  & \textbf{10.45} & 0.57 & 16.28(m) & \textbf{16.03} & 2.29  & 36(m)\\
\textbf{ERRL2(2opt)} & 6.18 & 0.65 & 5.49(m)  & 10.56 & 1.63  & 24.28(m) & 16.16 & 3.12 & 65(m)\\\hline
\end{tabular}
\label{tabile:randomdata}
\end{table*}

\subsection{Capacitated Vehicle Routing Problem (CVRP)}

In this table we group baselines as solver name as LKH3, non-learning baselines, constructive approaches \cite{nazari2018reinforcement} and \cite{kool2018attention} and improvement approaches (another two algorithms Chan and Tian~(\cite{chatting2018comparison}) and L2I~(\cite{lu2019learning}) introduced that fuses the strength of Operations Research (OR) heuristics with learning capabilities of reinforcement learning, in Table~\ref{tabile:randomdata}, we reported results from their paper). 

Our method is directly comparable to constructive approaches, given 1000 random CVRP instances of CVRP20, CVRO50 and CVRP100.  ERRL2(using \cite{nazari2018reinforcement} Policy network) slightly improve the solutions. However ERRL1(using \cite{kool2018attention} Policy network) find near optimal solutions for all the problem sizes. ERRL1 and ERRL2 combine with 2OPT outperforming all other learning approaches significantly both in terms of solution quality and solving time in Table~\ref{tabile:randomdata} all the results. For all results, the learning algorithms and baselines were implemented using their publicly available code except Chan and Tian~(\cite{chatting2018comparison}) and L2I~(\cite{lu2019learning}). 

\begin{figure}
  \begin{subfigure}[b]{0.45\columnwidth}
    \includegraphics[width=\linewidth]{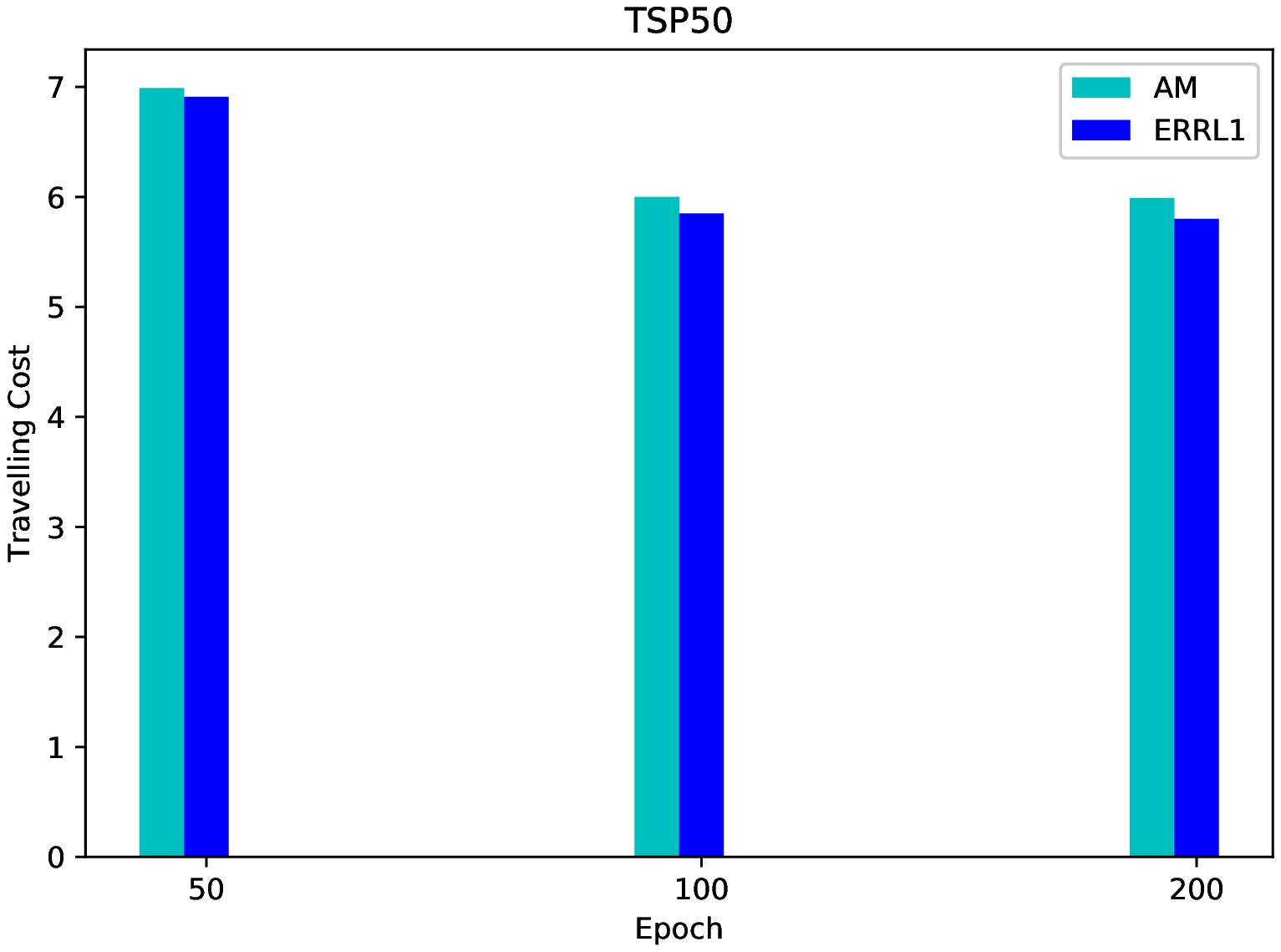}
    \caption{TSP Convergence}\label{fig:tsmalltime}
  \end{subfigure}
  \hfill 
  \begin{subfigure}[b]{0.45\columnwidth}
    \includegraphics[width=\linewidth]{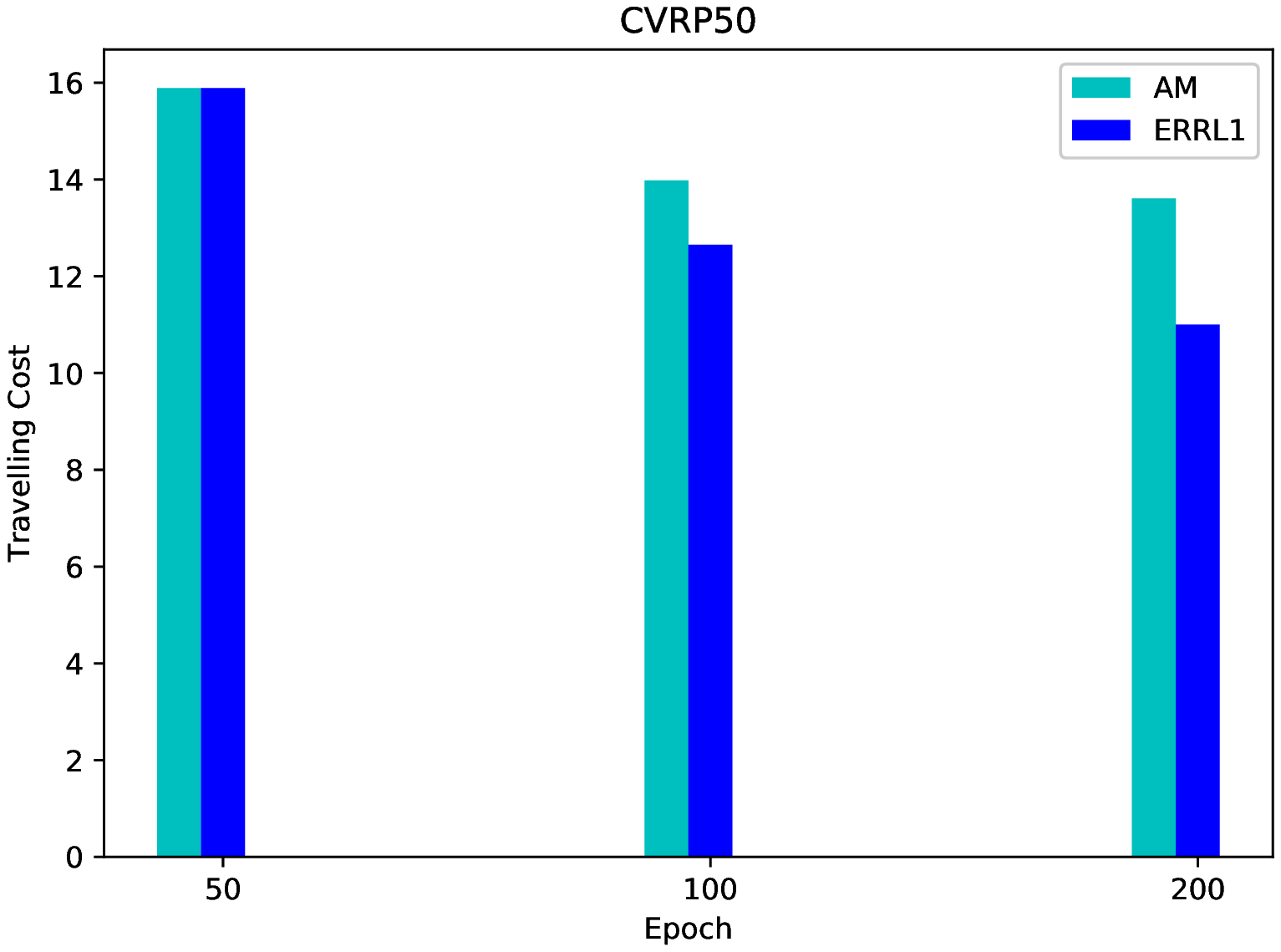}
    \caption{CVRP convergence}\label{fig:vsmalltime}
  \end{subfigure}
   \caption{Learning curves for TSP50 and CVRP50 compare to Kool et al \cite{kool2018attention} and our ERRL method on the same neural net (AM).}\label{fig:smalltsp}
\end{figure}

Learning curves of CVRP50 and TSP50 in Figure~\ref{fig:smalltsp} show that ERRL training is more stable and most of the learning converge faster than kool et al. \cite{kool2018attention} model. We observed also most of the learning is already completed within 200 epochs for both the problems in Figure~\ref{fig:smalltsp}. After each training epoch, we generate 1000 random instances to use them as a validation set.

\subsection{Multiple Routing with Fixed Fleet Problems (MRPFF)}

Analyse the  generalisation of ERRL method we created a new route problems to test how our model performs compare to other existing methods. The result for Multiple Routing with Fixed Fleet Problems (MRPFF) reported Table~\ref{tabile:mrpff}. MRPFF experiment of applying ERRL results with 20, 50, and 100 customer nodes are reported in Table~\ref{tabile:mrpff}, and all the ERRL models is shown to outperform all the existing methods. 

\begin{table*}[!ht]
\centering
\caption{Average tour length, gap percentage and the average solution time. Nazari et al.~\cite{nazari2018reinforcement} and Kool et al.~\cite{kool2018attention} we implemented and reported the results using their publicly available codes. The gap percentages reported with respect to optimal value. The running time reported in minutes (m) and seconds (s).}
\scriptsize
\begin{tabular}{|l|lll|lll|lll|}\hline
Method & \multicolumn{3}{|c|}{MRPFF=20} & \multicolumn{3}{|c|}{MRPFF=50}& \multicolumn{3}{|c|}{MRPFF=100} \\ \hline
\textbf{MRPFF}  & TourL  & Gap(\%) & Time(s) & TourL &  Gap(\%) &  Time(s)& TourL  &   Gap(\%) & Time(s)\\\hline

LKH3      & 5.34  & 0.00 & 17(m) & 9.12  & 0.00& 90(m) & 13.16 & 0.00 & 145(m)\\ \hline
\textbf{Constructive Models} &  &  & &  &  &  &  &  &  \\\hline
Nazari et al.\cite{nazari2018reinforcement} & 6.79 & 27.15  & 6 & 10.85 & 18.96 & 18.98  & 15.97 &21.35 &  33 \\
Kool et al.\cite{kool2018attention} & 5.99 & 12.17 & 0.01 & 10.30  & 12.93& 0.03  & 14.67 & 11.47 &  0.07 \\\hline
\textbf{Ours(Constructive)} &  &  & &  &  &    \\\hline
\textbf{ERRL1} & 5.51& 3.18& 0.01    & 10.10 & 10.74 & 0.03 & 13.97  & 6.15 & 0.07 \\
\textbf{ERRL2} &5.70 & 6.74 & 5    & 10.40  & 14.03  & 14 & 14.40 & 9.42  & 30 \\\hline
\textbf{ERRL1(2Opt)}  & \textbf{5.45} & 2.05& 2(m)  & \textbf{9.39} & 2.96 & 15(m)  & \textbf{13.80} & 4.86 &  23(m) \\
\textbf{ERRL2(2opt)} & {5.60} & 4.86  &  4.41(m) &{9.79} & 7.34  & 17(m) & \textbf{14.12} & 7.29 & 45(m) \\\hline
\end{tabular}
\label{tabile:mrpff}
\end{table*}

\subsection{Travelling Salesman Problem (TSP)}~\label{section:tsp} 

In this section, we want to evaluate and show our performance is comparable with existing work as the previous state-of-the-art approaches typically focus on TSP random data. For the TSP, we report optimal results by Concorde~(\cite{applegate2006traveling}) and~(\cite{helsgaun2017extension}). Besides, we compare against Nearest, Random and further Insertion, as well as Nearest Neighbour. In Table \ref{tabile:tsp} illustrates the performance of our techniques compared to the solver, heuristics, and state of the art learning techniques for various TSP instance sizes. Table \ref{tabile:tsp} is separated into four sections: solver, heuristics, learning methods using reinforcement learning (RL), and; learning models using supervised techniques (S). For all results were implemented using their publicly available code except Graph Convolutional Network (GCN) taken from~(\cite{joshi2019efficient}); We implemented  PN.,~(\cite{vinyals2015pointer}), Bello et al.,~(\cite{bello2016neural}), EAN.,~ (\cite{deudon2018learning}), ~[Kool~(\cite{kool2018attention}) and~[Nazari]~(\cite{nazari2018reinforcement}), accordingly refer the results we found from our implementation. We are able to achieve satisfactory results and report the average tour lengths of our approaches on TSP20, TSP50, and TSP100 in Table~\ref{tabile:tsp}. The data in Table \ref{tabile:tsp} shows~\textbf{[ERRL2]}, perform better compare to Nazari et al.~(\cite{nazari2018reinforcement}) for all sizes of TSP instances. The data in Table~\ref{tabile:tsp} shows using our greedy approach name Entropy regularised Reinforcement Learning~\textbf{[ERRL1]}, outperformed not only all the traditional baselines but also perform better compare to~~(\cite{kool2018attention}). 

Furthermore, we considered execution times in seconds for all instances. Run times are important but can vary due to implementation using Python or C++. We show the run times for our approach and compared with all the approaches, used python for implementation. Another important factor is using hardware such as GPUs or CPUs~(\cite{kool2018attention}). We implemented all approaches on the same hardware platform as experimental results can vary based on hardware platforms. In Table~\ref{tabile:tsp}, we report the running times for the results from our implementation using their publicly available codes, as reported by others, are not directly comparable. We only reported execution times for directly comparable baselines~(\cite{nazari2018reinforcement}) and~(\cite{kool2018attention}). We report the time it takes to solve the average solution time (in seconds) over a test set of size 1000 test.

\subsection{ERRL combine with 2-Opt}

In this study, another further enhancement is, we use a local search algorithm 2-opt~\cite{aarts2003local} to improve our results during test time. We show that the model can produce improve the result by using a ‘hybrid’ approach of a learned algorithm with local search. This hybrid approach is an example of combining learned and traditional heuristics. Recent many works showed that the design of the search procedure has an immense impact on the performance of the ML approach. Francois et al.~\cite{franccois2019evaluate} shown in their result that the search procedure can promote improvement, not from the learning intrinsic.  With local search added, the ERRL1 and ERRL2 with 2opt heuristics have much-improved performance than without 2-opt shown in Tables~\ref{tabile:randomdata},~\ref{tabile:mrpff} and~\ref{tabile:tsp}.

\begin{table*}[h]
\centering
\caption{Average tour length (TourL) and the gap percentages reported with respect to optimal value. * indicate that values are reported from~(\cite{joshi2019efficient}). EAN(2Opt)~\cite{deudon2018learning}, Nazari et al.~\cite{nazari2018reinforcement} and Kool et al.~\cite{kool2018attention} we implemented and reported the results using their publicly available codes. The gap percentages reported with respect to optimal value. The running time reported in minutes (m) and seconds (s).}
\scriptsize
\begin{tabular}{|l|lll|lll|lll|}\hline
\hline
Method & \multicolumn{3}{|c|}{TSP20} & \multicolumn{3}{|c|}{TSP50}& \multicolumn{3}{|c|}{TSP100} \\ \hline
\textbf{Solver}    & TourL  & Gap(\%)& Time & TourL  &  Gap(\%) & Time & TourL   & Gap(\%) & Time\\\hline
Concorde      & 3.83  & 0.00&  4m & 5.70  & 0.00& 10m & 7.77  & 0.00&55m\\
LKH3 & 3.83  & 0.00 &42(s) & 5.70 & 0.00 &59(m)& 7.77 &0.00&25(m)\\ \hline
\textbf{Heuristics} &  &  & &  &  &  &&&  \\\hline
Nearest Insertion &  4.33  & 13.05 & 1(s)& 6.78 & 18.94 &2(s)&  9.45 &21.62&6(s)\\ 
Random Insertion & 4.00   & 4.43 & 0(s) & 6.13 & 7.54 &1(s)& 8.51 &9.52&3(s)\\ 
Farthest Insertion & 3.92  & 2.34 & 1(s)& 6.01 & 5.43 &2(s)& 8.35  &7.46&7(s)\\ \hline
Or-tools &  3.85 & 0.52& -& 5.80  & 1.75 &-& 8.30  & 6.82& -\\\hline
\textbf{Learning Models (SL)} &  &&&&  & &  &  &     \\\hline
PN  & 3.88  & 1.30 && 6.62 & 16.14 & &  10.88 & 40.20&\\
GCN* & 3.86  & 0.78 &6(s)& 5.87   & 2.98&55(s) & 8.41 & 8.23& 6(m)\\\hline
\textbf{Learning Models (RL)} &  & &&& & &  &  &    \\\hline
Bello et al.\cite{bello2016neural} & 3.89 &  1.56&- & 5.99  & 5.08 &-& 9.68  &24.73&-\\
EAN.(2Opt)\cite{deudon2018learning}  & 3.93  & 2.61 & 4(m) & 6.63   & 16.31 &26(m)& 9.97 & 28.31&178(m)\\
Kool et all\cite{kool2018attention}& 3.85 &0.52 & 0.001(s)& {5.80}&1.75& 2(s)  & {8.15}  &4.89&6(s)\\
Nazari et al\cite{nazari2018reinforcement}& 4.00  & 4.43& & 7.01 & 22.76& & 9.46 &21.75&\\\hline
\textbf{Ours} &  &  & &  &  &  &&&  \\\hline
\textbf{ERRL1}  & \textbf{3.83} & 0& 0.01(s)&\textbf{5.74}  & 0.70& 2(s) & \textbf{7.86}  & 1.15 & 6(s)\\
\textbf{ERRL2} & 3.86 & 0.78 & 7(s) & {5.76} & 1.05& 18(s)& 7.92 &1.93 & 31(s)\\\hline
\textbf{ERRL1(2opt)}  & \textbf{3.81} & - & 1(m) & \textbf{5.71}  & 1.57& 11.87(m)& \textbf{7.77} & 0 & 25(m)\\ 
\textbf{ERRL2(2opt)} & 3.83 & 0 & 4(m) & {5.73} & 0.52&13(m) & 7.80 & 0.38 &37(m)\\
\hline
\end{tabular}
\label{tabile:tsp}
\end{table*}
\subsection{Solution search strategy Impact}

Recent work L2I  developed by \cite{lu2019learning} outperforms LKH3, our methods differ from L2I in terms of speed also ERRl is a purely data-driven way to solve COP. The ERRL model combines local search operator 2-opt but during test time. L2I is a specialised routing problem solver based on a handcrafted pool of improvement operators and perturbation operators. ERRL net is a construction type neural net models for CO problems, previous methods \cite{nazari2018reinforcement} \cite{kool2018attention} and including our method have two modes for inference in general. More inference techniques used during inference time such as greedy search, using greedy search a single deterministic trajectory is drawn using argmax on the policy. In “sampling mode,” multiple trajectories are sampled from the network following the probabilistic policy. 
\begin{figure}[!h]
  \vspace{-0.2cm}
  \centering
  \includegraphics[width=0.5\textwidth]{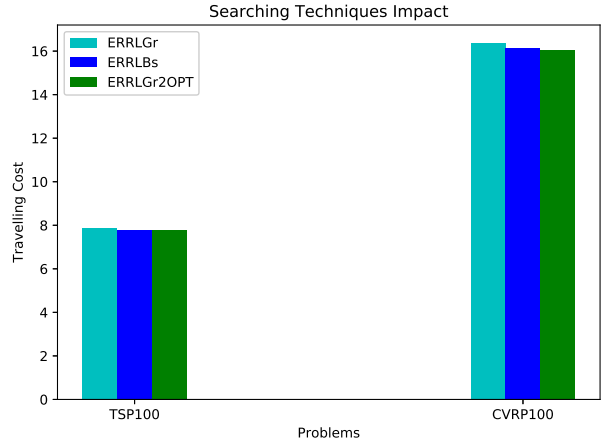}
  \caption{Comparison of Solution search setup with ERRL1}\label{fig:search}
  \vspace{-0.1cm}
\end{figure}

In the experiment, we used two different decoders: greedy describe in Figure~\ref{fig:search} referred to as name called [ERRL1Gr],  and the beam search (BS) in Figure~\ref{fig:search} as name [ERRL1BS]~(\cite{neubig2017neural}). Our results have shown that using the beam search algorithm, the quality of the solutions improved; however,  in computation time slightly increased. 

In this work, the two inference techniques and one inference technique combine with 2-opt operators during inference time experimentally showed that ML approaches benefited from a search procedure presented in Figure~\ref{fig:search}.  In Figure~\ref{fig:search}, greedy search, beam search combine with ERRL1 model, the optimality gap obtained 1.15\% and 0.38\% respectively, when we combine with 2 Opt operator with ERRL1 could be reduced from 1.15\% to 0 for TSPs of 100 nodes problems. However, the execution time is higher. We show the impact of a search procedure fuse with ML model performances in the figure~\ref{fig:search}. To better understand, we use greedy search, beam search and combined 2-opt with greedy search with ERRL1 to show that the importance of searching in ML-based approaches to combinatorial optimisation. 

\section{{Conclusion and Future Direction}}\label{sec:conclusion}

In this work, we presented the ERRL model that encourages exploration and finds model can improve performance on many optimisation problems incorporated with a policy gradient method. In the study, demonstrated that entropy augmented reward helps the model to avoid local optima and prevent premature convergence. The model has outperformed machine learning-based approaches significantly. We expect that the proposed architecture is not limited to route optimisation problems; it is an essential topic of future research to apply it to other combinatorial optimisation problems.

Recent learning-based algorithms for COP is Model-free deep RL methods that are notoriously expensive in terms of their sample complexity.  This challenge severely limits the applicability of model-free deep RL to real-world tasks. Our current ERRL-net model has less sample complexity. Nevertheless, it can be improved to use this model to real-world tasks, so model prevents brittleness concerning their hyper-parameters. Also, able to balance exploration and exploitation in terms of the task. Our future goal is to develop a model, instead of requiring the user to set the temperature manually (now we have done for alpha value), we can automate the process by reformulating a different maximum entropy reinforcement objective, where the entropy is treated as a constraint. Therefore, for future work, we need an effective method that can become competitive with the model-free state-of-the-art for combinatorial domains and more sample-efficient RL algorithms may be a key ingredient for learning from larger problems.  In addition, finding the most adapted search procedures for an ML model is still an open question.


\bibliographystyle{unsrt} 
\bibliography{my}
\newpage

\appendix
\section{Problems: TSP, CVRP, MRPFF}\label{apd:problems} 

The goal of our model is to generate the minimum total route length of the vehicle, where the route length need to be answerable from any routing problem, such as capacitated Vehicle Routing Problems(CVRP), multi-Vehicle Routing Problem with Fixed Fleet size (MRPFF) and TSP.
Let G (V; E) denote a weighted graph, where V is the set of nodes, E the set of edges. In an instance of the VRP problem, we are given a set of customer nodes, specifically, for VRP instance, the input (V = {$a_0$, . . . , $a_n$}) is a set of nodes, node $a_0$ represents the depot, and all other nodes represent the customers that need to be visited. There can be multiple vehicles serving customers. The Vehicle Routing Problem is to find a set of routes (all starting and ending at the depot) with minimal cost, and each customer must be visited by exactly one vehicle. Consider a set of customers, to these serve customers; we must design routes for each vehicle available, all starting from a single depot, $a_0$. Each route must start at the depot, visit a subset of customers and then return to that depot. The objective of the problem is to determine a set of minimal cost routes that satisfy all requirements defined above. With these parameters, the formulation of CVRP is given by, 
\[ 
min \sum_{\substack{
a_i, a_j \in V 
}} C_{a_ia_j}x_{a_ia_j}
\label{equation:mrp}
 \]

subject to, \\
$\displaystyle\sum_{a_i \in V} x_{a_ia_j}  = 1$   $\forall_{a_j} \in V  \setminus  \{a_0\}$,\\
$\displaystyle\sum_{a_j \in V} x_{a_ia_j}  = 1$  $\forall_{a_i} \in V \setminus  \{a_0\}$, \\
$\displaystyle\sum_{v_i \in V} x_{a_ia_o}  = K$, \\
$\displaystyle\sum_{v_j \in V} x_{a_oa_j}  = K$,\\
             $\displaystyle\sum_{i \not\in S} \displaystyle\sum_{j \in S} x_{a_ia_j}  \geq  r(S)$,
             where $\forall  S \subseteq V \setminus \{a_{0}\},  S\neq \emptyset$,  

In this formulation ${\displaystyle C_{a_ia_j}}$ represents the cost of going from node ${\displaystyle a_i}$  to node ${\displaystyle a_j}$ and, where the cost of traveling from node $a_i$ to $a_j$ is $c_{a_ia_j}\in \mathcal{R}^{+}$. ${\displaystyle x_{a_ia_j}}$ is a binary variable, $x_{a_ia_j} \in  \{0,1 \}$ and $a_i, a_j \in V$, that has value 1 if the edge going from ${\displaystyle a_i}$  to ${\displaystyle a_j}$ is considered as part of the solution and ${\displaystyle 0}$ otherwise, K is the number of available vehicles. r(S) is the minimum number of the vehicle to serve set S, the capacity cut constraints, which impose that the routes must be connected and that the demand on each route must not exceed the vehicle capacity. Also assuming that ${\displaystyle 0}$ is the depot node~(\cite{toth2002models}). In this work, we also use another variance of VRP, the multiple Routing problem with fixed fleet size (MRPFF), where customer locations are considered on a 2D Euclidean space, to serve customers, and we must design routes for one vehicle available at a single depot, $a_o$. Each route must start at the depot, visit a subset of customers and then return to that depot. It is further assumed that vehicle does not need to attain any demands but need to optimise the set of routes. The vehicles must create routes starting and ending at a depot node to optimise routes. The MRPFF has no demand, and we consider only one fixed vehicle, visited two sets of customers (two sets of routes). Another route problem is the TSP problem; we are given a set of points/cities. A tour of these cities is a sequence where each city is visited and only visited once. Then the TSP problem is to find such a tour of cities such that the total travel distance between consecutive pairs of cities in the tour is minimised.

\subsection{2-opt Local Search}

In this work as inference techniques, we use 2-opt to further improve the solutions. In a 2-opt algorithm, when removing two edges, there is only one alternative feasible solution. The procedure searches for k edge swaps that will be replaced by a new edge,  swapping techniques results in a shorter tour. Moreover, sequential pairwise operators such as k-opt moves can be decomposed in simpler l-opt ones, where l < k. For instance, in our work, 2-opt sequential operations decomposed into one \cite{helsgaun2009general}. 

\section{Policy Networks(ERRL1)} \label{apd:am} 

We used the neural architecture of~(\cite{kool2018attention}) to apply our approach to improve solutions of routing problems. Here, we briefly describe the attention model architecture~\cite{kool2018attention} in terms of the~\textbf{ERRL1)}. The model is consists of attention based encoder and decoder network. The encoder produces embeddings of all inputs and decoder produces the sequence~$\pi$ of given inputs, one at a time: encoder inputs the encoder embeddings and a problem specific mask and context. When partial tour constructed, it cannot be changed, and the rest of the nodes find a path from the last node to the first node and decoder network consists of embeddings of first and last node. 

The attention-based encoder embeds the input nodes and processed N sequential layers, each consisting multi-head attention and feed-forward sub-layer. The graph embedding is computed the node embeddings. The attention layer in this model following Transformer Architecture~(\cite{vaswani2017attention}), each attention layer has two sub-layers, one multi-head attention processes message passing between the nodes and another layer is a node wise fully connected feed-forward layer. The decoder is decoding the solutions sequentially at each time step. The decoder outputs the node $\pi_t$ based on the embeddings from the encoder. They also augmented the graph with a special context node to represent decoding context similar to~(\cite{kool2018attention}). Similar to~(\cite{kool2018attention}) computed output probabilities, add one decoder layer with a single attention head. We used our approach using~(\cite{kool2018attention}) model and reported improved results on a number of combinatorial optimisation (routing) problems.


\section{Neural Network Architecture(ERRL2)} \label{apd:nazari} 
\subsection{Policy Network} 

The Nazari et al.~\cite{nazari2018reinforcement} model used in the ERRL2 experiments as the same as Nazari et al.\cite{nazari2018reinforcement} for TSP and CVRP both the problems. For MRPFF we consider the problems setting as CVRP except there is no customer demand(which we refer to as “the original Nazari et al. paper~\cite{nazari2018reinforcement}”).

The policy network as a sequence to sequence (S2S) learning employed with an attention mechanism. S2S learning technique, sequentially given input to make a decision at each time step and generates the solutions as a sequence of customer locations. The customer locations are on a 2D Euclidean space. Given an input sequence, the model finds the conditional probability of the output sequence similar to~(\cite{sutskever2014sequence}). Commonly, recurrent neural networks are used in sequence-to-sequence models to estimate this conditional probability. The sequence-to-sequence model assumes elements of the output sequence is fixed. Unlike the sequence-to-sequence model, the VRP solution (output) is a permutation of the problem nodes (input). To achieve the solution, use an attention mechanism (see, for example,~(\cite{vinyals2015pointer}). The attention mechanism query information from all elements in the input nodes set. An affinity function is evaluated to assemble the output sequence (with each node and the final output of the model) to generate a set of scalars (aggregates many signals into one). Later, the softmax function applied to these scalars to obtain the attention weights given to each element of the input set at each time step.

We define the combinatorial optimisation problem with a given set of inputs. Between every decoding step, some of the elements of each input to change. For instance, in the case of VRP, the rest of the customer demands change over time as the vehicle visits the customer nodes; or we might consider a variant in which new customers arrive or adjust their demand values over time, independent of the vehicle decisions~(\cite{nazari2018reinforcement}). We formally, represent each input by a sequence of tuples. We start from an arbitrary input in, where we use the pointer to refer to that input and every decoding step will points to one available input, which regulates the input of the next decoder step until a terminating condition satisfied as Nazari et al.\cite{nazari2018reinforcement}. These inputs are given to an encoder which embeds into latent space vectors. These embedded vectors are combined with the output of a decoder. This points to one of the elements of the input. This process generates a sequence and ends when a terminating condition is satisfied, e.g., when a specific number of steps are completed. In dynamic route optimisation, for example, in case of CVRP, includes all customer locations as well as their demands, and the depot location; then, the remaining demands are updated with respect to the vehicle destination and its load, and the terminating condition is that there is no more demand to satisfy. This process will generate a sequence of length, possibly with a different sequence length compared to the input length.  For example, the vehicle may have to go back to the depot several times to refill. We are interested in finding a stochastic policy~$\pi$, which generates the sequence in a way that minimises a loss objective while satisfying the problem constraints. 

\subsection{Training [ERRL2]}\label{apd:first}

Algorithm \textbf{ERRL}~\ref{algo:ERRLLNet}, we have two networks with weight vectors $\theta$ and $\phi$ associated with actor and critic networks, respectively. Summarises, we parameterise the stochastic policy~$\theta$ with parameters~$\phi$. Policy gradient methods improve the policy iteratively use an estimate of the gradient of the expected return with respect to the policy parameters. Let us consider a family of problems, denoted by~$I$, and a probability distribution over them, denoted by~$\phi_I$. We draw~$N$ sample problems from~$I$ and use Monte Carlo simulation to produce feasible sequences for the current policy $\pi_{\theta}$. In Algorithm~\ref{algo:ERRLLNet}, we taken the variables of the nth instance is referring to as the superscript of~$n$. RL objective is to learn the parameters of some policy such that the expected sum of rewards is maximised under the induced trajectory distribution. After termination of the decoding in all~$N$ problems, we calculate the corresponding rewards in step 13. During training in step 14 computed the policy gradient as
\begin{equation}
\begin{matrix}
     \nabla\theta_{\pi} = \dfrac{1}{N} \sum_{n=1}^{N} (R^{n} - V(s^n_0; \phi))\nabla_{\theta} log P(a^n|s^n_0)
   \label{equation:pg2}
\end{matrix}
\end{equation}
We add entropy bonus to avoid premature convergence and reduce variance in the gradient in step 14. In step 14,
~$V(s^n_0,\phi)$ is the reward approximation for problem instance ~$n$. We also update the critic network~$\nabla\theta_{\phi}$ in the direction of reducing the difference between the expected rewards with the Monte Carlo estimation of the reward (calculated from the critic network) in step 15. The complete training procedure in Algorithm~\ref{algo:ERRLLNet}.

\begin{algorithm}
\footnotesize
\SetKwFunction{This}{this}


Procedure Training (Policy network :$\pi_{\theta}$:; $\phi$:  random weights for Critic network\, training set S );
\For{$iteration = 1,2, \cdots$}{

reset gradients $d\theta \leftarrow 0, d\phi \leftarrow 0$

sample N problems according to $\phi_{I}$
   
 \For{$n = 1$ to $N$}{
initialise step count  $t\leftarrow 0$
    
      \Repeat{}{
     select $a^n_{t+1}$ according to the distribution $P(a^n_{t+1}|a^n_{t}, s^n_{t})$
     
     observe new state $a^n_{t+1}$
     
    $t \leftarrow t+1$
     
} termination condition is satisfied
     
compute reward $R^n = R(a^n, s^n_0)$
  
}
  $\nabla\theta_{\pi} = \dfrac{1}{N} \sum_{n=1}^{N} (R^{n} - V(s^n_0; \phi))\nabla_{\theta} log P(a^n|S^n_0) +\alpha  H (\pi_{\theta} (. |s_{t}))$
  
$\nabla\theta_{\phi} = \dfrac{1}{N}  \sum_{n=1}^{N} \nabla_{\phi} (R^{n} - V(s^n_0; \phi))^2$

update $\theta_{\pi}$ using  $\nabla\theta_{\pi}$ and $\phi$ using $\nabla\theta_{\phi}$
}
\caption{ERRL Algorithm}\label{algo:ERRLLNet}
\end{algorithm}


\subsection{Datasets and Settings}\label{apd:dataset}

In this work, we implemented the datasets described by~(\cite{nazari2018reinforcement}). The locations and demands are randomly generated from a fixed distribution. Specifically, the customers and depot locations are randomly generated in the unit square, and the demand of each node chosen randomly uniform and is a discrete number in~$\{1\cdots9\}$, chosen randomly uniform. We note, the demand values can be generated from any distribution. 
For faster training and generating feasible solutions, we have used a masking scheme which sets the log-probabilities of infeasible solutions to~(~-$\infty$) or forces a solution if the condition is satisfied. In the CVRP, we use the masking procedures for nodes with zero demand are not allowed to be visited; all customer nodes will be masked if the vehicle’s remaining load is exactly 0, and the customers whose demands are greater than the current vehicle load are masked. Under this masking scheme, the vehicle must satisfy all customer’s demands when visiting it. 

We used the same architecture settings throughout all the experiments and datasets. Across all experiments, we use one-dimensional convolutional operation, LSTM cells with 128 hidden units. For training both networks, we use the REINFORCE Algorithm and Adam optimiser~(\cite{kingma2014adam}) with a learning rate of~$0.0001$. Similarly~(\cite{nazari2018reinforcement}) the decay rate of every 5000 steps by a factor of 0.96. In the critic network, first, we use the output probabilities of the actor-network to compute a weighted sum of the embedded inputs, and then, it has two hidden layers: one dense layer with ReLU activation and another linear one with a single output. The variables in both actor and critic network are initialised with Xavier initialisation~(\cite{glorot2010understanding}). The batch size N is 128, dropout with probability 0.1 in the decoder LSTM, and we clip the gradients when their norm is greater than 2. In the experiment, we show having an entropy augmented reward and, in general, a more stochastic policy changes this objective and perform better in terms of speed of learning shown in Section~\ref{sec:exp}. We have manually set entropy values for all the problems. As we achieve best performance for ERRL1 for value $\alpha=0.3$, we used the same hyper parameters value in ERRL2 model.

\section{Set of parameters Testing}\label{section:parameter} 
In this section, We trained the model using different set of parameters and illustrates the result in Tables \ref{tabile:net}, \ref{tabile:net1} and \ref{tabile:net3} for problem size 20, 50 and 100 respectively on CVRP dataset. We evaluated our model with changing parameters of the model. In this experiment we first trained our model with VRP50 node dataset and tested on VRP20, VRP50 and VRP100 instances.
\begin{table*}[h]
\small
\centering
\caption{ In this table, we reported average tour length (TourL) for CVRP20 using various set of parameters.}
\begin{tabular}{|l|l|l|l|l|l|l|l|l|l|l|l|l|}
\hline
Co-efficient &0.5&0.6&0.7&0.8&0.9\\\hline
Learning Rate &&&&&\\\hline
0.00001  &7.60 & &   & &  \\ \hline
0.0001  &8.22&&   & &  \\ \hline
0.00001 &&7.10&   & &  \\ \hline
0.0001  &&7.56&   & & \\ \hline
0.00001  &&& 7.26  & &  \\ \hline
0.0001   &&&  7.85 & & \\ \hline
0.00001  &&&   & 7.80  &  \\ \hline
0.0001   &&&   &  6.99 & \\ \hline
0.00001  &&&   &  &  6.38\\ \hline
0.0001   &&&   &  & 7.05 \\ \hline
\end{tabular}
\label{tabile:net}
\end{table*}
 
\begin{table*}[h]
\small
\centering
\caption{ In this table, we reported average tour length (TourL) for CVRP50 using various set of parameters.}
\begin{tabular}{|l|l|l|l|l|l|l|l|l|l|l|l|l|}
\hline
Co-efficient &0.5&0.6&0.7&0.8&0.9\\\hline
Learning Rate&&&&&\\\hline
0.00001  &13.13 & &   & &  \\ \hline
0.0001  &12.19&&   & &  \\ \hline
0.00001  &&12.21&   & &  \\ \hline
0.0001    &&11.19&   & & \\ \hline
0.00001  &&& 12.70  & &  \\ \hline
0.0001   &&& 11.58  & & \\ \hline
0.00001  &&&   & 11.56 &  \\ \hline
0.0001   &&&   & 12.29  & \\ \hline
0.00001  &&&   &  &  10.95\\ \hline
0.0001   &&&   &  &12.91 \\ \hline
\end{tabular}
\label{tabile:net1}
\end{table*}
\begin{table*}[h]
\small
\centering
\caption{ In this table, we reported average tour length (TourL) for CVRP100 using various set of parameters.}
\begin{tabular}{|l|l|l|l|l|l|l|l|l|l|l|l|l|}
\hline
Co-efficient &0.5&0.6&0.7&0.8&0.9\\\hline
Learning Rate &&&&&\\\hline
0.00001 & 17.57 & &   & &  \\ \hline
0.0001 & 17.35&&   & &  \\ \hline
0.00001  &&16.12 &   & &  \\ \hline
0.0001    &&20.85&   & & \\ \hline
0.00001  &&& 18.03  & &  \\ \hline
0.0001   &&& 15.33  & & \\ \hline
0.00001  &&&   &  24.12 &  \\ \hline
0.0001  &&&   & 19.87  & \\ \hline
0.00001  &&&   &  & 17.51 \\ \hline
0.0001   &&&   &  & 19.94\\ \hline
\end{tabular}
\label{tabile:net3}
\end{table*}

\end{document}